\documentclass[11pt,a4paper]{article}
\usepackage{acl2014}
\usepackage{times}
\usepackage{url}
\usepackage{latexsym}
\usepackage{graphicx}
\usepackage{amsmath}
\usepackage{mathtools}
\usepackage{color}
\usepackage{multirow}
\usepackage{algorithmicx}
\usepackage{algorithm}
\usepackage{algpascal}
\usepackage{algc}
\usepackage{algcompatible}
\usepackage{algpseudocode}
\usepackage{breqn}
\usepackage{float}
\usepackage{caption}
\usepackage{anyfontsize}
\usepackage{verbatim}

\DeclareMathOperator*{\argmax}{arg\,max}

\DeclareMathOperator*{\avgSim}{avgSim}
\DeclareMathOperator*{\avgSimC}{avgSimC}
\DeclareMathOperator*{\globalSim}{globalSim}
\DeclareMathOperator*{\localSim}{localSim}

\title{Efficient  Non-parametric Estimation of \\ Multiple Embeddings per Word in Vector Space}

\author{Arvind Neelakantan\textsuperscript{*}, Jeevan Shankar\textsuperscript{*}, Alexandre Passos, Andrew McCallum  \\
  Department of Computer Science \\
University of Massachusetts, Amherst \\ Amherst, MA, 01003 \\
  {\tt \{arvind,jshankar,apassos,mccallum\}@cs.umass.edu} \\
 }

\date{}
\begin{document}

\maketitle
\begin{abstract}
\let\thefootnote\relax\footnote{\textsuperscript{*}The first two authors contributed equally to this paper.}
There is rising interest in vector-space word embeddings and their use
in NLP, especially given recent methods for their fast estimation at
very large scale.  Nearly all this work, however, assumes a single
vector per word type---ignoring polysemy and thus jeopardizing their
usefulness for downstream tasks.  We present an extension to the
Skip-gram model that efficiently learns multiple embeddings per word
type.  It differs from recent related work by jointly performing word
sense discrimination and embedding learning, by non-parametrically
estimating the number of senses per word type, and by its efficiency and
scalability.  We present new state-of-the-art results in the  word similarity in context task and
demonstrate its scalability by training with one machine on a corpus
of nearly 1 billion tokens in less than 6 hours.
\end{abstract}

\section{Introduction}

Representing words by dense, real-valued vector embeddings, also
commonly called ``distributed representations,'' helps address the
curse of dimensionality and improve generalization because they can
place near each other words having similar semantic and syntactic
roles.  This has been shown dramatically in state-of-the-art results
on language modeling \cite{bengio:2003,mnih:2007} as well as
improvements in other natural language processing tasks
\cite{collobert:2008,turian:2010}.  Substantial benefit arises when
embeddings can be trained on large volumes of data.  Hence the recent
considerable interest in the CBOW and Skip-gram models of
\newcite{mikolov:2013a}; \newcite{mikolov:2013b}---relatively simple log-linear
models that can be trained to produce high-quality word embeddings on
the entirety of English Wikipedia text in less than half a day on one
machine.

There is rising enthusiasm for applying these models to improve
accuracy in natural language processing, much like Brown clusters
\cite{brown:1992} have become common input features for many tasks,
such as named entity extraction \cite{miller:2004,ratinov:2009} and
parsing \cite{koo:2008,tackstrom:2012}.  In comparison to Brown
clusters, the vector embeddings have the advantages of substantially
better scalability in their training, and intriguing potential for
their continuous and multi-dimensional interrelations.  In fact,
\newcite{passos:2014} present new state-of-the-art results in CoNLL
2003 named entity extraction by directly inputting continuous vector
embeddings obtained by a version of Skip-gram that injects supervision
with lexicons.  Similarly \newcite{bansal:2014} show results in
dependency parsing using Skip-gram embeddings.  They have also
recently been applied to machine translation
\cite{zou:2013,mikolov:2013c}.

A notable deficiency in this prior work is that each word type ({\it
  e.g.} the word string {\sf\small plant}) has only one vector
representation---polysemy and hononymy are ignored.  This results in
the word {\sf\small plant} having an embedding that is approximately the
average of its different contextual semantics relating to biology,
placement, manufacturing and power generation.  In moderately
high-dimensional spaces a vector can be relatively ``close'' to
multiple regions at a time, but this does not negate the unfortunate
influence of the triangle inequality\footnote{For distance $d$, $d(a,c) \leq d(a,b) + d(b,c)$.} here: words that are
not synonyms but are synonymous with different senses of the same word
will be pulled together.  For example, {\sf\small pollen} and {\sf\small refinery}
will be inappropriately pulled to a distance not more than the sum of
the distances {\sf\small plant--pollen} and {\sf\small plant--refinery}.  Fitting
the constraints of legitimate continuous gradations of semantics are
challenge enough without the additional encumbrance of these
illegitimate triangle inequalities.




Discovering embeddings for multiple senses per word type is the focus
of work by \newcite{reisinger:2010b} and \newcite{huang:2012}.  They
both pre-cluster the contexts of a word type's tokens into
discriminated senses, use the clusters to re-label the corpus' tokens
according to sense, and then learn embeddings for these re-labeled
words.  The second paper improves upon the first by employing an
earlier pass of non-discriminated embedding learning to obtain vectors
used to represent the contexts.  Note that by pre-clustering, these
methods lose the opportunity to jointly learn the sense-discriminated
vectors and the clustering.  Other weaknesses include their fixed
number of sense per word type, and the computational expense of the
two-step process---the \newcite{huang:2012} method took one week of
computation to learn multiple embeddings for a 6,000 subset of the
100,000 vocabulary on a corpus containing close to billion tokens.\footnote{Personal
  communication with authors Eric H. Huang and Richard Socher.}


This paper presents a new method for learning vector-space embeddings
for multiple senses per word type, designed to provide several
advantages over previous approaches.  (1) Sense-discriminated vectors
are learned jointly with the assignment of token contexts to senses;
thus we can use the emerging sense representation to more accurately
perform the clustering.  (2) A non-parametric variant of our method
automatically discovers a varying number of senses per word type.  (3)
Efficient online joint training makes it fast and scalable.
We refer to our method as {\it Multiple-sense Skip-gram}, or {\it
  MSSG}, and its non-parametric counterpart as {\it NP-MSSG}.


Our method builds on the Skip-gram model \cite{mikolov:2013a}, but
maintains multiple vectors per word type.  During online training with a
particular token, we use the average of its context words' vectors to
select the token's sense that is closest, and perform a gradient
update on that sense. In the non-parametric version of our method,
we build on {\it facility location} \cite{Meyerson:2001}: a new cluster is
created with probability proportional to the distance from the context
to the nearest sense.

We present experimental results demonstrating the benefits of our
approach.  We show qualitative improvements over single-sense
Skip-gram and \newcite{huang:2012}, comparing against word neighbors from
our parametric and non-parametric methods.  We present quantitative
results in three tasks.  On both the SCWS and WordSim353 data sets our
methods surpass the previous state-of-the-art.  The Google Analogy
task is not especially well-suited for word-sense evaluation since its
lack of context makes selecting the sense difficult; however
our method dramatically outperforms \newcite{huang:2012} on this task.
Finally we also demonstrate scalabilty, learning multiple senses,
training on nearly a billion tokens in less than 6 hours---a 27x improvement on
Huang et al.

\section{Related Work}
\label{sec:related-work}

Much prior work has focused on learning vector representations of
words; here we will describe only those most relevant to
understanding this paper. Our work is based on neural language models,
proposed by \newcite{bengio:2003}, which extend the traditional idea
of $n$-gram language models by replacing the conditional probability
table with a neural network, representing each word token by a small
vector instead of an indicator variable, and estimating the parameters
of the neural network and these vectors jointly. Since the
\newcite{bengio:2003} model is quite expensive to train, much research
has focused on optimizing it. \newcite{collobert:2008} replaces the
max-likelihood character of the model with a max-margin approach,
where the network is encouraged to score the correct $n$-grams higher
than randomly chosen incorrect $n$-grams. \newcite{mnih:2007} replaces
the global normalization of the Bengio model with a tree-structured
probability distribution, and also considers multiple positions for
each word in the tree. 

More relevantly, \newcite{mikolov:2013a} and
\newcite{mikolov:2013b} propose extremely computationally efficient
log-linear neural language models by removing the hidden layers of the
neural networks and training from larger context windows with very
aggressive subsampling. The goal of the models in
\newcite{mikolov:2013a} and \newcite{mikolov:2013b} is not so much
obtaining a low-perplexity language model as learning word
representations which will be useful in downstream tasks. Neural
networks or log-linear models also do not appear to be necessary to
learn high-quality word embeddings, as \newcite{dhillon:2011} estimate
word vector representations using Canonical Correlation Analysis
(CCA). 

Word vector representations or embeddings have been used in various
NLP tasks such as named entity recognition
\cite{neelakantan:2014,passos:2014,turian:2010}, dependency parsing
\cite{bansal:2014}, chunking \cite{turian:2010,dhillon:2011},
sentiment analysis \cite{mass:2011}, paraphrase detection
\cite{socher:2011} and learning representations of paragraphs and
documents \cite{le:2014}. The word clusters obtained from Brown
clustering \cite{brown:1992} have similarly been used as features in
named entity recognition \cite{miller:2004,ratinov:2009} and
dependency parsing \cite{koo:2008}, among other tasks.

There is considerably less prior work on learning multiple vector
representations for the same word type. \newcite{reisinger:2010b}
introduce a method for constructing multiple sparse, high-dimensional
vector representations of words. \newcite{huang:2012} extends this approach incorporating global document context to learn
multiple dense, low-dimensional embeddings by using recursive neural
networks.  Both the methods perform word sense discrimination as a
pre-processing step by clustering contexts for each word type, making
training more expensive.  While methods such as those described in
\newcite{dhillon:2011} and \newcite{reddy:2011} use token-specific
representations of words as part of the learning algorithm, the final
outputs are still one-to-one mappings between word types and word
embeddings.


\section{Background: Skip-gram model}
The Skip-gram model learns word embeddings such that they are useful in predicting the surrounding words in a sentence. In the Skip-gram model, $v(w) \in R^{d}$ is the vector representation
of the word $w \in W$, where $W$ is the words vocabulary and $d$ is
the embedding dimensionality.
 
Given a pair of words $(w_t,c)$, the probability that the word $c$ is
observed in the context of word $w_t$ is given by,
\begin{equation}
P(D = 1 | v(w_t), v(c)) = \frac{1}{1 + e^{-v(w_t)^Tv(c)} }
\end{equation}
The probability of not observing word $c$ in the context of $w_t$ is given by, 
\begin{align*}
P(D = 0 | v(w_t), v(c)) &= \\ & 1 - P(D = 1 | v(w_t), v(c))
\end{align*}

Given a training set containing the sequence of word types $w_1, w_2,
\ldots, w_{T}$, the word embeddings are learned by maximizing the
following objective function:
\begin{align*}
J(\theta) &= \sum_{(w_{t}, c_{t}) \in D^{+}}{ \sum_{c \in c_{t}} \log  P(D = 1 | v(w_t), v(c)) } \\
      & + \sum_{(w_{t}, c'_{t}) \in D^{-}}{ \sum_{c' \in c'_{t}} \log P(D = 0 | v(w_t), v(c'))}
\end{align*}
where $w_{t}$ is the $t^{th}$ word in the training set, $c_{t}$ is the
set of observed context words of word $w_{t}$ and $c'_{t}$ is the set
of randomly sampled, noisy context words for the word $w_{t}$. $D^{+}$
consists of the set of all observed word-context pairs $(w_{t},
c_{t})$ ($t=1,2\ldots,T$). $D^{-}$ consists of pairs $(w_{t}, c'_{t})$
($t=1,2\ldots,T$) where $c'_{t}$ is the set of randomly sampled, noisy
context words for the word $w_{t}$.

\begin{figure}
\includegraphics[scale=0.43]{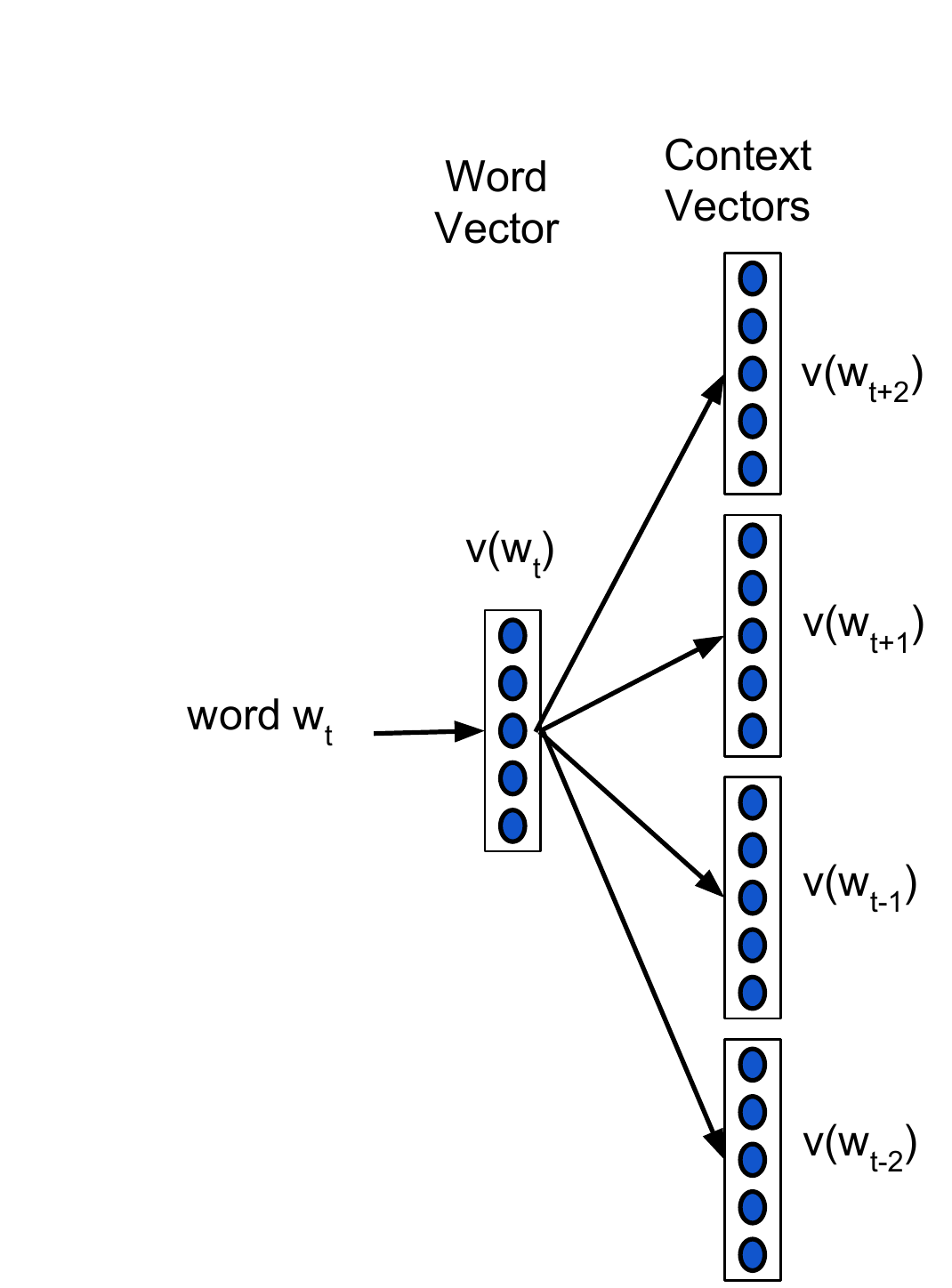}
\caption{Architecture of the Skip-gram model with window size $R_{t}=2$.  Context $c_t$ of word $w_t$  consists of $w_{t-1}, w_{t-2}, w_{t+1}, w_{t+2}$.}
\label{fig1}
\end{figure}

For each training word $w_t$, the set of context words $c_{t} = \{ w_{t-R_{t}}, \ldots, w_{t-1}, w_{t+1}, \ldots, w_{t+R_{t}} \}$ includes $R_{t}$ words to the left and right of the given word as shown in Figure \ref{fig1}. $R_{t}$ is the window size considered for the word $w_t$ uniformly randomly sampled from the set $\{1, 2, \ldots, N\}$, where N is the maximum context window size.

The set of noisy context words $c'_{t}$ for the word $w_{t}$ is constructed by randomly sampling  $S$ noisy context words for each word in the context $c_t$.  The noisy context words are randomly sampled from the following distribution, 
\begin{equation}
P(w) = \frac{p_{unigram}(w)^{3/4}}{Z} 
\end{equation}
where $p_{unigram}(w)$ is the unigram distribution of the words and $Z$ is the normalization constant.

\section{Multi-Sense Skip-gram (MSSG) model}
To extend the Skip-gram model to learn multiple embeddings per word  we follow previous work \cite{huang:2012,reisinger:2010b} and let each sense of word have its own embedding, and  induce the senses by clustering the embeddings of the context words around each token. The vector representation of the context is the average of its context words' vectors. For every word type, we maintain clusters of its contexts and the sense of a word token is predicted as the cluster that is closest to its context representation. After predicting the sense of a word token, we perform a gradient update on the embedding of that sense. The crucial difference from previous approaches is that word sense discrimination and learning embeddings are performed jointly by predicting the sense of the word using the current parameter estimates.

\begin{figure}
\includegraphics[scale=0.4]{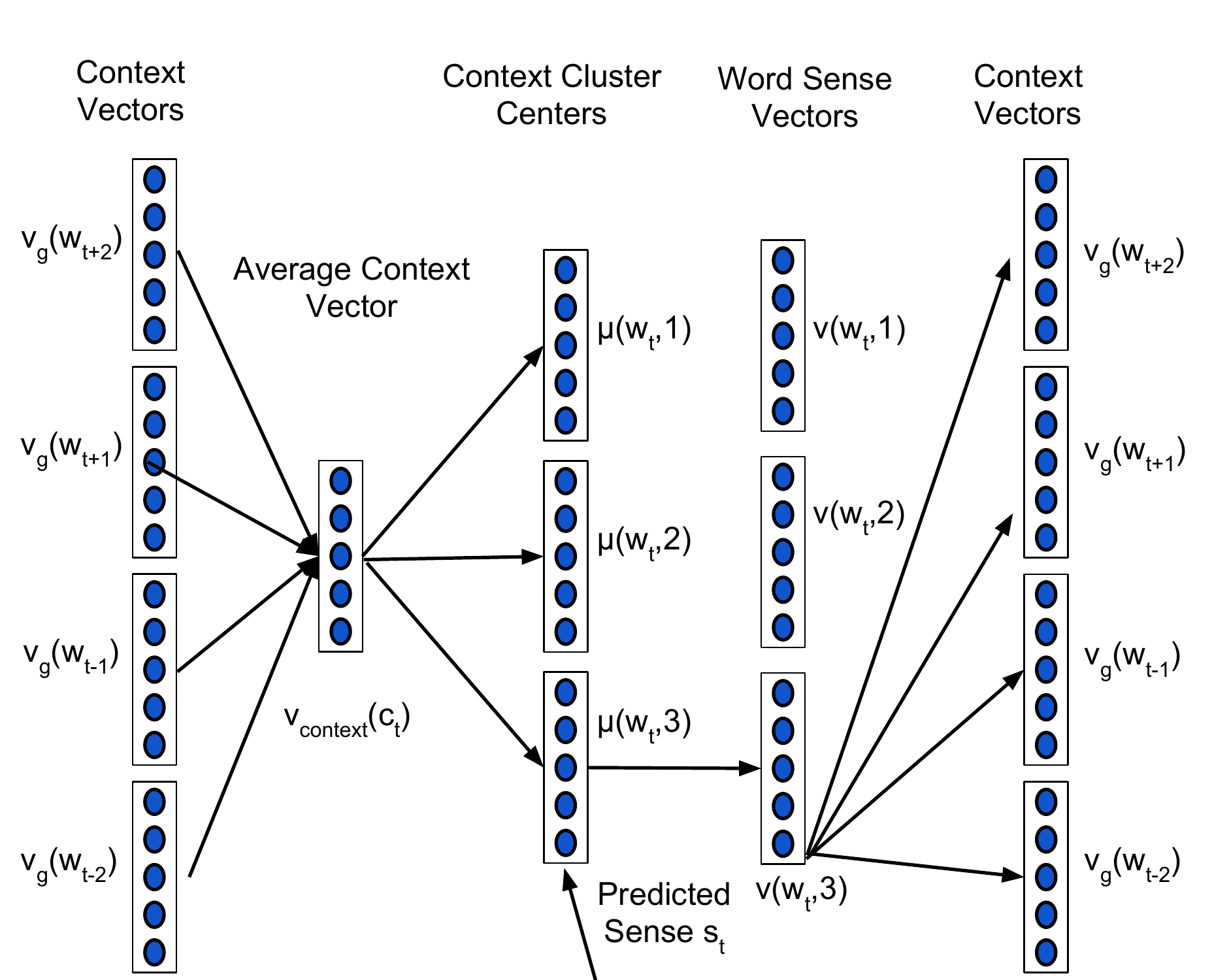}
\caption{Architecture of Multi-Sense Skip-gram (MSSG) model with window size $R_{t}=2$ and $K=3$.  Context $c_t$ of word $w_t$  consists of $w_{t-1}, w_{t-2}, w_{t+1}, w_{t+2}$. The sense is predicted by finding the cluster center of the context that is closest to the average of the context vectors.}
\label{fig2}
\end{figure}

In the MSSG model, each word $w \in W$ is associated with a global vector $v_g(w)$ and each sense of the word has an embedding (sense vector) $v_{s}(w,k)$ ($k=1, 2, \ldots, K$) and  a context cluster with center $\mu(w,k)$ ($k=1, 2, \ldots, K$). The $K$ sense vectors and the global vectors are of dimension $d$ and $K$ is a hyperparameter.

Consider the word $w_t$ and let $c_t=\{ w_{t-R_{t}}, \ldots, w_{t-1}, w_{t+1}, \ldots, w_{t+R_{t}} \}$ be the set of observed context words. The vector representation of the context is defined as the average of the global vector representation of the words in the context. Let $v_{context}(c_t) = \frac{1}{2*R_{t}} \sum_{c \in c_{t} }v_g(c)$ be the vector representation of the context $c_t$. We use the global vectors of the context words instead of its sense vectors to avoid the computational complexity associated with predicting the sense of the context words. We predict $s_{t}$, the sense of word $w_{t}$ when observed with context $c_{t}$ as the context cluster membership of the vector $v_{context}(c_t)$ as shown in Figure \ref{fig2}. More formally, 
\begin{equation}
s_t = \argmax_{k=1,2,\ldots,K}  sim(\mu(w_t, k), v_{context}(c_t))
\end{equation}
The hard cluster assignment is similar to the $k$-means algorithm. The
cluster center is the average of the vector representations of all the
contexts which belong to that cluster.  For $sim$ we use cosine similarity in our experiments.

Here, the probability that the word $c$ is observed in the context of word $w_t$ given the sense of the word $w_{t}$ is,
\begin{align*}
P(D = 1 | s_t, &v_s(w_t, 1), \ldots, v_s(w_t, K), v_g(c)) \\ &=P(D = 1 | v_s(w_t, s_t), v_g(c)) \\&= \frac{1}{1 + e^{-v_s(w_t, s_{t})^Tv_g(c)} }
\end{align*}
The probability of not observing word $c$ in the context of $w_t$ given the sense of the word $w_{t}$ is, 
\begin{align*}
P(D = 0 | s_t, &v_s(w_t, 1), \ldots, v_s(w_t, K), v_g(c))  \\ &=P(D = 0 | v_s(w_t, s_t), v_g(c))  \\&= 1 - P(D = 1 |v_s(w_t, s_t), v_g(c))
\end{align*}

Given a training set containing the sequence of word types $w_1, w_2, ..., w_T$, the word embeddings are learned by maximizing the following objective function:
\begin{align*}
&J(\theta)= \\ & \sum_{(w_t, c_t) \in D^{+}}{ \sum_{c \in c_{t}}  \log P(D = 1| v_s(w_t, s_t), v_g(c) )  } + \\
      & \sum_{(w_t, c'_t) \in D^{-}}{ \sum_{c' \in c'_{t}}  \log P(D = 0 | v_s(w_t, s_t), v_g(c')) }
\end{align*}
where $w_t$ is the $t^{th}$ word in the sequence, $c_t$ is the set of observed context words and $c'_t$ is the set of noisy context words for the word $w_t$. $D^{+}$ and $D^{-}$ are constructed in the same way as in the Skip-gram model. 

After predicting the sense of word $w_t$, we update the embedding of the predicted sense for the word $w_t$ ($v_s(w_t, s_t)$), the global vector of the words in the context and the global vector of the randomly sampled, noisy context words. The context cluster center of cluster $s_t$ for the word $w_t$ ($\mu(w_t, s_t)$) is updated since context $c_t$ is added to the cluster $s_t$.

\newfloat{algorithm}{t}{lop}
\begin{algorithm}
\caption{Training Algorithm of MSSG model}
\begin{algorithmic}[1]
 \State Input: $w_1, w_2, ..., w_T$, $d$, $K$, $N$.
 \State Initialize $v_{s}(w,k)$ and $v_g(w)$, $\forall w \in W, k \in \{1, \ldots, K\}$ randomly, $ \mu(w, k)$ $\forall w \in W, k \in \{1, \ldots, K\}$ to 0.
 \For{$t = 1, 2, \ldots, T$ }
 	\State $R_t \sim \{1,\ldots,N\}$
 	\State $c_t=\{w_{t-R_{t}},\ldots,w_{t-1},w_{t+1},\ldots,w_{t+R_{t}}\}$
 	\State $v_{context}(c_t) = \frac{1}{2*R_{t}} \sum_{c \in c_{t} }v_g(c)$
 	\State $s_t = \argmax_{k=1,2,\ldots,K}$ \{  \par 
 	\hskip\algorithmicindent $ sim(\mu(w_t, k), v_{context}(c_t))$\} \par
 	\State Update context cluster center $\mu(w_t, s_t)$ since context $c_t$ is added to context cluster $s_t$ of word $w_t$. 
 	\State $c'_t = Noisy\_Samples(c_t)$		 		 			
	\State Gradient update on $v_s(w_t, s_t)$, global vectors of words in $c_t$ and $c'_t$.	 
 \EndFor
\State Output: $v_{s}(w,k)$, $v_g(w)$ and context cluster centers $\mu(w,k)$, $\forall w \in W, k \in \{1, \ldots, K\}$

\end{algorithmic}
\end{algorithm}

\section{Non-Parametric MSSG model (NP-MSSG)}

The MSSG model learns a fixed number of senses per word type. In this
section, we describe a non-parametric version of MSSG, the
NP-MSSG model, which learns varying number of senses per word type. Our
approach is closely related to the online non-parametric clustering
procedure described in \newcite{Meyerson:2001}. We create a new
cluster (sense) for a word type with probability proportional to the
distance of its context to the nearest cluster (sense).

Each word $w \in W$ is associated with sense vectors, context clusters
and a global vector $v_g(w)$ as in the MSSG model. The number of senses
for a word is unknown and is learned during training. Initially, the
words do not have sense vectors and context clusters. We create the
first sense vector and context cluster for each word on its first
occurrence in the training data. After creating the first context
cluster for a word, a new context cluster and a sense vector are
created online during training when the word is observed with a
context were the similarity between the vector representation of the
context with every existing cluster center of the word is less than
$\lambda$, where $\lambda$ is a hyperparameter of the model.

Consider the word $w_t$ and let $c_t=\{ w_{t-R_{t}}, \ldots, w_{t-1}, w_{t+1}, \ldots, w_{t+R_{t}} \}$ be the set of observed context words. The vector representation of the context is defined as the average of the global vector representation of the words in the context. Let $v_{context}(c_t) = \frac{1}{2*R_{t}} \sum_{c \in c_{t} }v_g(c)$ be the vector representation of the context $c_t$. Let $k(w_t)$ be the  number of context clusters or the number of senses  currently associated with word $w_t$. $s_t$, the sense of word $w_t$ when $k(w_t) > 0$ is given by
\begin{equation}
    s_t =
    \begin{cases}
      k(w_t) + 1,  & \text{if}   \max_{k=1,2,\ldots,k(w_t)} \{  
       sim\\ & ( \mu(w_t, k),  v_{context}(c_t)) \} < \lambda \\
      k_{max}, & \text{otherwise}
    \end{cases}
  \end{equation}
where $\mu(w_t, k)$ is the cluster center of the $k^{th}$ cluster of word $w_t$ and $k_{max} = \argmax_{k=1,2,\ldots,k(w_t)} sim(\mu(w_t, k), v_{context}(c_t))$. 

The cluster center is the average of the vector representations of all the contexts which belong to that cluster. If $s_t = k(w_t) + 1$, a new context cluster and a new sense vector are created for the word $w_t$.
	
The NP-MSSG model and the MSSG model described previously differ only in the way word sense discrimination is performed. The objective function and the probabilistic model associated with observing a (word, context) pair given the sense of the word remain the same.

\section{Experiments}
\label{sec:experiments}

To evaluate our algorithms we train embeddings using the same corpus
and vocabulary as used in \newcite{huang:2012}, which is the April
2010 snapshot of the Wikipedia corpus \cite{wiki-corpus-2010}.  It
contains approximately 2 million articles and 990 million tokens.  In
all our experiments we remove all the words with less than 20
occurrences and use a maximum context window ($N$) of length 5 (5
words before and after the word occurrence). We fix the number of
senses ($K$) to be 3 for the MSSG model unless otherwise specified.
Our hyperparameter values were selected by a small amount of manual
exploration on a validation set.  In NP-MSSG we set $\lambda$ to -0.5.
The Skip-gram model, MSSG and NP-MSSG models sample one noisy context
word ($S$) for each of the observed context words. We train our models
using AdaGrad stochastic gradient decent \cite{duchi:2011} with
initial learning rate set to 0.025. Similarly to \newcite{huang:2012},
we don't use a regularization penalty.

Below we describe qualitative results, displaying the embeddings and
the nearest neighbors of each word sense, and quantitative experiments
in two benchmark word similarity tasks.

\begin{table}
\centering
\begin{tabular}{|l|r|}
\hline 
Model & Time (in hours) \\ \hline \hline
Huang et al & 168 \\ \hline
MSSG 50d  & 1  \\ \hline
MSSG-300d & 6 \\ \hline
NP-MSSG-50d & 1.83  \\ \hline
NP-MSSG-300d &  5 \\ \hline \hline
Skip-gram-50d &  0.33\\ \hline
Skip-gram-300d &  1.5  \\ \hline
\end{tabular}
\caption{Training Time Results. First five model reported in the table are capable of learning multiple embeddings for each word and Skip-gram  is capable of learning only single embedding for each word.}
\label{table:training-time-exp}
\end{table}

Table \ref{table:training-time-exp} shows time to train our models,
compared with other models from previous work. All these times are
from single-machine implementations running on similar-sized
corpora. We see that our model shows significant improvement in the
training time over the model in \newcite{huang:2012}, being within
well within an order-of-magnitude of the training time for Skip-gram models.

\begin{table}[t!]
{\fontsize{7.6}{9.12} \selectfont
\begin{tabular}{|l|l|}

\multicolumn{2}{l}{\textsc{Apple}}\\ \hline
 Skip-gram  & blackberry,  macintosh, acorn, pear, plum  \\  \hline
 \multirow{3}{*}{MSSG} & pear, honey, pumpkin, potato, nut \\
 &  microsoft, activision, sony, retail,  gamestop \\ 
 &  macintosh,  pc, ibm, iigs, chipsets\\ \hline
 \multirow{2}{*}{NP-MSSG} &  apricot, blackberry, cabbage, blackberries, pear \\ 
 & microsoft, ibm,   wordperfect, amiga, trs-80\\
 \hline 

\multicolumn{2}{l}{\textsc{Fox}}\\ \hline
  Skip-gram & abc, nbc, soapnet, espn, kttv \\  \hline
 \multirow{3}{*}{MSSG} & beaver, wolf, moose, otter, swan \\
 & nbc, espn, cbs, ctv, pbs \\ 
 & dexter, myers, sawyer, kelly, griffith \\ 
 \hline
 \multirow{2}{*}{NP-MSSG} & rabbit, squirrel, wolf, badger, stoat \\
 & cbs,abc, nbc, wnyw, abc-tv \\ \hline
 
 \multicolumn{2}{l}{\textsc{Net}}\\ \hline
 Skip-gram  & profit, dividends, pegged, profits, nets \\  \hline
 \multirow{3}{*}{MSSG} & snap, sideline, ball, game-trying, scoring  \\
 & negative, offset, constant, hence, potential \\ 
 & pre-tax, billion, revenue, annualized, us\$ \\ \hline 
 \multirow{4}{*}{NP-MSSG} & negative, total, transfer, minimizes, loop \\ 
 & pre-tax, taxable, per, billion, us\$, income \\ 
 & ball, yard, fouled, bounced, 50-yard \\
 & wnet, tvontorio, cable, tv, tv-5 \\
 \hline 
 
 \multicolumn{2}{l}{\textsc{Rock}}\\ \hline
 Skip-gram  & glam, indie, punk, band, pop \\  \hline
 \multirow{3}{*}{MSSG} & rocks, basalt, boulders, sand, quartzite \\ 
 & alternative, progressive, roll, indie, blues-rock \\
 & rocks, pine, rocky, butte, deer \\ \hline
 \multirow{2}{*}{NP-MSSG} & granite, basalt, outcropping, rocks, quartzite \\ 
 & alternative, indie, pop/rock, rock/metal, blues-rock \\
\hline 
 
\multicolumn{2}{l}{\textsc{Run}}\\ \hline
 Skip-gram  & running, ran, runs, afoul, amok  \\  \hline
 \multirow{3}{*}{MSSG}& running, stretch, ran, pinch-hit, runs  \\
 & operated , running, runs, operate, managed \\
 & running, runs, operate, drivers, configure \\ \hline
 \multirow{4}{*}{NP-MSSG} &  two-run, walk-off, runs, three-runs, starts \\
 & operated, runs, serviced, links, walk \\
 & running, operating, ran, go, configure \\
 & re-election, reelection, re-elect, unseat, term-limited \\
 & helmed, longest-running, mtv, promoted, produced \\
 \hline
 
 \end{tabular}
 }
 \caption{Nearest neighbors of each sense of each word, by cosine similarity, for different algorithms. Note that the different senses closely correspond to intuitions regarding the senses of the given word types.}
\label{table:knn}
\end{table}

\subsection{Nearest Neighbors}
\label{sec:qual-analys-near}

Table~\ref{table:knn} shows qualitatively the results of discovering
multiple senses by presenting the nearest neighbors associated with
various embeddings.  The nearest neighbors of a word are computed by
comparing the cosine similarity between the embedding for each sense
of the word and the context embeddings of all other words in the
vocabulary.  Note that each of the discovered senses are indeed
semantically coherent, and that a reasonable number of senses are
created by the non-parametric method. Table \ref{table:knn-with-socher} shows the nearest neighbors of the word plant for Skip-gram, MSSG , NP-MSSG and Haung's model \cite{huang:2012}. 
\begin{table}[t!]
\small
\begin{tabular}{|p{0.7cm}|p{6.4cm}|}
 \hline
 Skip-gram  & plants, flowering, weed, fungus, biomass  \\  \hline
 \multirow{3}{*}{\parbox[t]{0.7cm}{MS\\-SG} } & plants, tubers, soil, seed, biomass \\
 & refinery, reactor, coal-fired, factory, smelter \\ 
 & asteraceae, fabaceae, arecaceae, lamiaceae, ericaceae\\ \hline
 \multirow{4}{*}{\parbox[t]{0.7cm}{NP\\MS\\-SG}} & plants, seeds,  pollen, fungal, fungus \\
 & factory,  manufacturing,  refinery, bottling, steel \\
 & fabaceae, legume, asteraceae, apiaceae, flowering  \\
 & power, coal-fired, hydro-power, hydroelectric, refinery \\ \hline
\multirow{10}{*}{\parbox[t]{0.7cm}{Hua\\-ng et al \\} } & insect, capable, food, solanaceous, subsurface \\ 
 & robust, belong, pitcher, comprises, eagles \\
 & food, animal, catching, catch, ecology, fly  \\
 & seafood, equipment, oil, dairy, manufacturer\\
 & facility,  expansion, corporation, camp, co. \\
 & treatment, skin, mechanism, sugar, drug \\
 & facility, theater, platform, structure, storage \\
 & natural, blast, energy, hurl, power \\
 & matter, physical, certain, expression, agents  \\
 & vine, mute, chalcedony, quandong,  excrete\\
\hline
\end{tabular}
\caption{Nearest Neighbors of the word {\it plant} for different models. We see that the discovered senses in both our models are more semantically coherent than \newcite{huang:2012} and NP-MSSG is able to learn reasonable number of senses.}
\label{table:knn-with-socher}
\end{table}

\subsection{Word Similarity}
\label{sec:word-similarity}

We evaluate our embeddings on two related datasets: the WordSim-353
\cite{finkelstein:2001} dataset and the Contextual Word Similarities
(SCWS) dataset \newcite{huang:2012}.

WordSim-353 is a standard dataset for evaluating word vector
representations. It consists of a list of pairs of word types, the
similarity of which is rated in an integral scale from 1 to 10. Pairs
include both monosemic and polysemic words. These scores to each word
pairs are given without any contextual information, which makes them
tricky to interpret. 

To overcome this issue, Stanford's Contextual Word Similarities (SCWS)
dataset was developed by \newcite{huang:2012}. The dataset consists of
2003 word pairs and their sentential contexts. It consists of 1328
noun-noun pairs, 399 verb-verb pairs, 140 verb-noun, 97
adjective-adjective, 30 noun-adjective, 9 verb-adjective, and 241
same-word pairs. We evaluate and compare our embeddings on both
WordSim-353 and SCWS word similarity corpus.

Since it is not trivial to deal with multiple embeddings per word, we
consider the following similarity measures between words $w$ and $w'$
given their respective contexts $c$ and $c'$, where $P(w, c, k)$ is
the probability that $w$ takes the $k^{th}$ sense given the context
$c$, and $d(v_s(w, i), v_s(w', j))$ is the similarity measure between
the given embeddings $v_s(w, i)$ and $v_s(w', j)$.

The $\avgSim$ metric,
\begin{dmath*}
\avgSim(w, w') = 
    \frac{1}{K^{2}} \sum_{i=1}^{K} \sum_{j=1}^{K} d\left(v_s(w, i), v_s(w', j)\right),
\end{dmath*}
computes the average similarity over all embeddings for each word, ignoring
information from the context.

To address this, the $\avgSimC$ metric,
\begin{align*}
\avgSimC(w,w') = 
    \sum_{j=1}^{K} & \sum_{i=1}^{K} P(w,c, i) P(w',c',j) \\ & \times d\left(v_s(w, i), v_s(w', j)\right)
\end{align*}
weighs the similarity between each pair of senses by how well does
each sense fit the context at hand.

The $\globalSim$ metric uses each word's global context
vector, ignoring the many senses:
\begin{dmath*}
\globalSim(w, w') = 
    d\left(v_g(w), v_g(w')\right).
\end{dmath*}

Finally, $\localSim$ metric selects a single sense for each word based
independently on its context and computes the similarity by
\begin{dmath*}
\localSim(w, w') = 
  d\left(v_s(w, k), v_s(w', k')\right),
\end{dmath*}
where $k = \argmax_{i} P(w, c, i)$ and $k' = \argmax_{j} P(w', c', j)$ and $P(w, c, i)$ is the probability that $w$ takes the $i^{th}$ sense given context $c$. The probability of being in a cluster is calculated as the inverse of the cosine distance to the cluster center \cite{huang:2012}.

We report the Spearman correlation between a model's similarity scores
and the human judgements in the datasets.

\begin{table*}[t!]
\centering
\begin{tabular}{ |l|r|r|r|r| }
\hline
Model & $\globalSim$ & $\avgSim$ & $\avgSimC$ & $\localSim$ \\  \hline \hline
TF-IDF                        & 26.3 & - & - & - \\ \hline
Collobort \& Weston-50d       & 57.0  & - & - & -   \\ \hline
Skip-gram-50d            &  63.4     & - & -  & - \\ \hline
Skip-gram-300d           &  65.2 & - & - & - \\ \hline
\hline
Pruned TF-IDF                 & 62.5 & 60.4 & 60.5 & -   \\ \hline
Huang et al-50d               & 58.6 & 62.8 & 65.7 & 26.1 \\ \hline
MSSG-50d          & 62.1 & 64.2 & 66.9 &  49.17 \\ \hline 
MSSG-300d     & 65.3 & 67.2 & \textbf{69.3} & 57.26 \\ \hline
NP-MSSG-50d    & 62.3 & 64.0 & 66.1 & 50.27 \\ \hline 
NP-MSSG-300d & \textbf{65.5} & \textbf{67.3} & 69.1 & \textbf{59.80}   \\ \hline 
\end{tabular}
\caption{Experimental results in the SCWS task. The numbers are Spearman’s correlation $\rho \times 100$ between each model's similarity judgments and the human judgments, in context. First three models learn only a single embedding per model and hence, $\avgSim$, $\avgSimC$ and $\localSim$ are not reported for these models, as they'd be identical to $\globalSim$. Both our parametric and non-parametric models outperform the baseline models, and our best model achieves a score of 69.3 in this task. NP-MSSG achieves the best results when $\globalSim$, $\avgSim$ and $\localSim$ similarity measures are used. The best results according to each metric are in bold face.}
\label{table:scws-task}
\end{table*}

\begin{table}
\centering
\begin{tabular}{|l|r|}
\hline 
Model & $\rho \times 100 $ \\ \hline \hline
HLBL & 33.2 \\ \hline
C\&W & 55.3 \\ \hline
Skip-gram-300d & 70.4 \\ \hline
Huang et al-G & 22.8 \\ \hline
Huang et al-M & 64.2 \\ \hline
MSSG 50d-G & 60.6 \\ \hline
MSSG 50d-M & 63.2 \\ \hline
MSSG 300d-G& 69.2 \\ \hline
MSSG 300d-M& \textbf{70.9} \\ \hline  
NP-MSSG 50d-G & 61.5 \\ \hline
NP-MSSG 50d-M & 62.4 \\ \hline
NP-MSSG 300d-G& 69.1 \\ \hline
NP-MSSG 300d-M& 68.6 \\ \hline  \hline
Pruned TF-IDF & 73.4 \\ \hline 
ESA & 75 \\ \hline
Tiered TF-IDF & 76.9 \\ \hline
\end{tabular}
\caption{Results on the WordSim-353 dataset. The table shows the Spearman’s correlation $\rho$ between the model's similarities and human judgments.  G indicates the $\globalSim$ similarity measure and M indicates $\avgSim$ measure.The best results among models  that learn low-dimensional and dense representations are in bold face. Pruned TF-IDF \cite{reisinger:2010b}, ESA \cite{gabrilovich:2007} and Tiered TF-IDF \cite{reisinger:2010a} construct spare, high-dimensional representations.}
\label{table:ws353-task}
\end{table}

Table \ref{table:ws353-task} shows the results on WordSim-353
task. C\&W refers to the language model by \newcite{collobert:2008}
and HLBL model is the method described in \newcite{mnih:2007}.  On
WordSim-353 task, we see that our model performs significantly better
than the previous neural network model for learning multi-representations
per word \cite{huang:2012}. Among the methods that learn
low-dimensional and dense representations, our model performs slightly
better than Skip-gram.
Table \ref{table:scws-task} shows the results for the SCWS task. In
this task, when the words are given with their context, our model
achieves new state-of-the-art results on SCWS as shown in the
Table-\ref{table:scws-task}. The previous state-of-art model
\cite{huang:2012} on this task achieves $65.7\%$ using the $\avgSimC$
measure, while the MSSG model achieves the best score of $69.3\%$ on this
task. The results on the other metrics are similar. For a fixed embedding dimension, the model by \newcite{huang:2012} has more parameters than our model since it uses a hidden layer. The results show that our model performs better than \newcite{huang:2012} even when both the models use 50 dimensional vectors and the performance of our model improves as we increase the number of dimensions to 300.  
\begin{figure}
\includegraphics[scale=0.3]{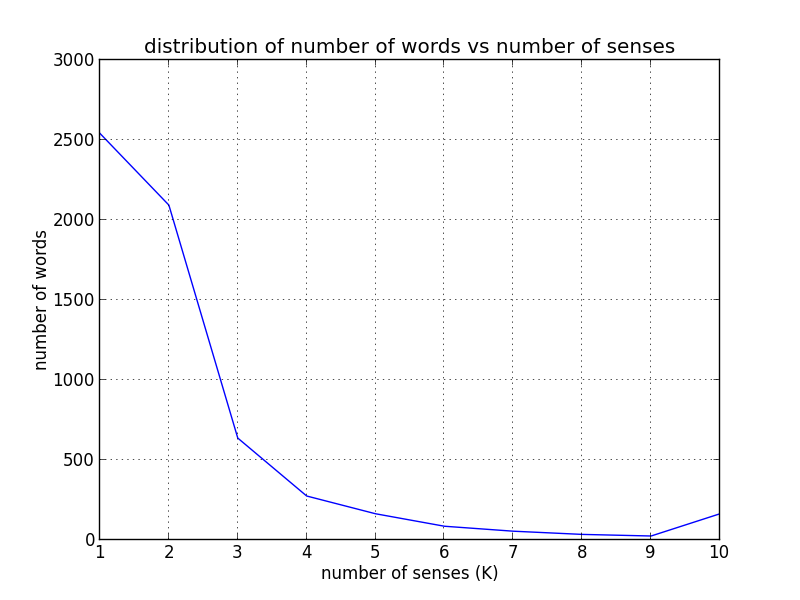}
\label{fig:np-mssg-sense-dist} 
\caption{The plot shows the distribution of number of senses learned per word type in NP-MSSG model}
\end{figure}

\begin{figure*}
\centering
\includegraphics[width=\textwidth]{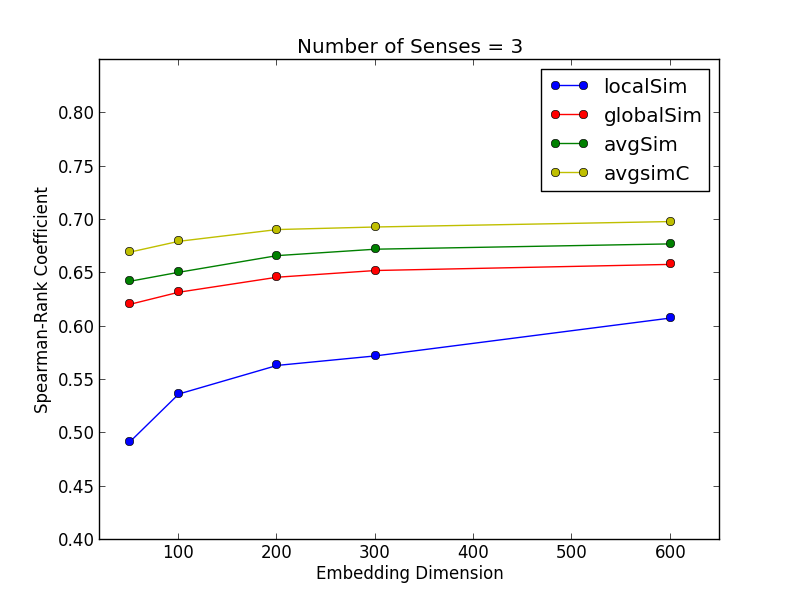}
\label{fig:sense} 
\caption{Shows the effect of varying embedding dimensionality of the MSSG Model on the SCWS task.}
\end{figure*}
\begin{figure*}
\includegraphics[width=\textwidth]{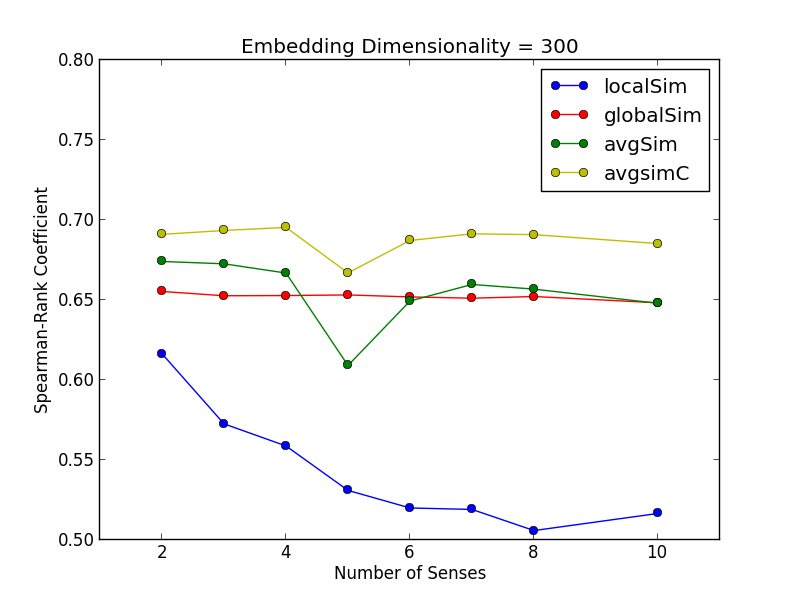}
\label{fig:sense} 
\caption{show the effect of varying number of senses of the MSSG Model on the SCWS task.}
\end{figure*}


We evaluate the models in a word analogy task introduced by
\newcite{mikolov:2013a} where both MSSG and NP-MSSG models achieve 64\% accuracy compared to
12\% accuracy by \newcite{huang:2012}. Skip-gram which is the state-of-art model for this task achieves 67\% accuracy.
\begin{table}
\centering
\begin{tabular}{|l|r|r|r|}
\hline
Model & Task & $Sim$ & $\rho \times 100$ \\ \hline
Skip-gram & WS-353 & $\globalSim$ & 70.4 \\ \hline
MSSG & WS-353 & $\globalSim$ & 68.4 \\ \hline
MSSG& WS-353 & $\avgSim$ & \textbf{71.2} \\  \hline
NP MSSG & WS-353 & $\globalSim$ & 68.3\\ \hline
NP MSSG & WS-353 & $\avgSim$ &  69.66 \\ \hline \hline
MSSG & SCWS & $\localSim$  & 59.3 \\ \hline
MSSG & SCWS & $\globalSim$ &  64.7\\ \hline 
MSSG & SCWS & $\avgSim$ & 67.2 \\ \hline
MSSG & SCWS & $\avgSimC$ & \textbf{69.2} \\ \hline
NP MSSG & SCWS & $\localSim$ & 60.11\\ \hline
NP MSSG & SCWS & $\globalSim$ & 65.3 \\ \hline
NP MSSG & SCWS & $\avgSim$ & 67 \\ \hline
NP MSSG & SCWS & $\avgSimC$ & 68.6 \\ \hline
\end{tabular}
\caption{Experiment results on WordSim-353 and SCWS Task. Multiple Embeddings are learned for top 30,000 most frequent words in the vocabulary. The embedding dimension size is 300 for all the models for this task. The number of senses for MSSG model is 3.}
\label{table:30k-word-sim-task}
\end{table}

Figure 3 shows the distribution of number of senses learned
per word type in the  NP-MSSG model.  
We learn the multiple embeddings
for the same set of approximately 6000 words that were used in
\newcite{huang:2012} for all our experiments to ensure fair
comparision. These approximately 6000 words were choosen by Huang et al.\ mainly
from the top 30,00 frequent words in the vocabulary.  This selection
was likely made to avoid the noise of learning multiple senses for infrequent words. However, our
method is robust to noise, which can be seen by the good performance of
our model that learns multiple embeddings for the top 30,000 most
frequent words.  We found that even by learning multiple embeddings
for the top 30,000 most frequent words in the vocubulary, MSSG model
still achieves state-of-art result on SCWS task with an $\avgSimC$
score of 69.2 as shown in Table \ref{table:30k-word-sim-task}.

\section{Conclusion} 
We present an extension to the Skip-gram model that efficiently learns multiple embeddings per word type. The model jointly performs word sense discrimination and embedding learning, and  non-parametrically estimates the number of senses per word type. Our method achieves new state-of-the-art results in the word similarity in context task 
and learns multiple senses, training on close to billion tokens in
less than 6 hours. The global vectors, sense vectors and cluster centers of
our model and  code for learning them are available at \url{https://people.cs.umass.edu/~arvind/emnlp2014wordvectors}.
In future work we plan to use the
multiple embeddings per word type in downstream NLP tasks.

\section*{Acknowledgments}
This work was supported in part by the Center for Intelligent Information Retrieval and in part by DARPA under agreement number FA8750-13-2-0020. The U.S. Government is authorized to reproduce and distribute reprints for Governmental purposes notwithstanding any copyright notation thereon. Any opinions, findings and conclusions or recommendations expressed in this material are those of the authors and do not necessarily reflect those of the sponsor.



\end{document}